\documentclass[times,twocolumn,final]{elsarticle}

\usepackage{prletters}
\usepackage{framed,multirow}

\usepackage{amssymb}
\usepackage{latexsym}

\usepackage{url}
\usepackage{xcolor}
\definecolor{newcolor}{rgb}{.8,.349,.1}

\usepackage{subcaption}
\usepackage{booktabs}
\usepackage{lipsum}
\usepackage{paralist}
\usepackage{scrextend} 
\usepackage{stfloats}
\usepackage{array}
\usepackage{blindtext}
\usepackage{enumitem}
\usepackage{amsmath}

\journal{Pattern Recognition Letters}

\begin{document}

\ifpreprint
  \setcounter{page}{1}
\else
  \setcounter{page}{1}
\fi

\begin{frontmatter}
\title{Neural sentence embedding using only in-domain sentences for out-of-domain sentence detection in dialog systems}

\author[1]{Seonghan \snm{Ryu}\corref{cor1}}
\ead{ryush@postech.ac.kr}
\cortext[cor1]{Corresponding author: 
  Tel.: +82-54-279-5567;  
  fax: +82-54-279-2299;}
\author[2]{Seokhwan \snm{Kim}} 
\ead{kims@i2r.a-star.edu.sg}
\author[1]{Junhwi \snm{Choi}}
\ead{chasunee@postech.ac.kr}
\author[1]{Hwanjo \snm{Yu}}
\ead{hwanjoyu@postech.ac.kr}
\author[1]{Gary Geunbae \snm{Lee}}
\ead{gblee@postech.ac.kr}

\address[1]{Pohang University of Science and Technology (POSTECH), 77 Cheongam-Ro, Nam-Gu, Pohang, 37673, Republic of Korea}
\address[2]{Institute for Infocomm Research (I2R), 1 Fusionopolis Way, \# 21-01 Connexis (South Tower), 138632, Singapore}

\received{1 May 2013}
\finalform{10 May 2013}
\accepted{13 May 2013}
\availableonline{15 May 2013}
\communicated{}

\begin{abstract}

To ensure satisfactory user experience, dialog systems must be able to determine whether an input sentence is \textit{in-domain} (ID) or \textit{out-of-domain} (OOD).
We assume that only ID sentences are available as training data because collecting enough OOD sentences in an unbiased way is a laborious and time-consuming job.
This paper proposes a novel neural sentence embedding method that represents sentences in a low-dimensional continuous vector space that emphasizes aspects that distinguish ID cases from OOD cases.
We first used a large set of unlabeled text to pre-train \textit{word representations} that are used to initialize neural sentence embedding.
Then we used domain-category analysis as an auxiliary task to train neural sentence embedding for OOD sentence detection.
After the sentence representations were learned, we used them to train an autoencoder aimed at OOD sentence detection.
We evaluated our method by experimentally comparing it to the state-of-the-art methods in an eight-domain dialog system; our proposed method achieved the highest accuracy in all tests.
\end{abstract}


\end{frontmatter}



\section{Introduction}
\label{sec:intro}

Dialog systems provide natural-language interfaces between humans and machines.
Because human conversation can range among topics, many studies have been recently conducted on multi-domain dialog systems~\citep{Hakkani-Tur:Interspeech2016, Jiang:IWSDS2014, Lee:TASLP2013, Ryu:IWSDS2015, Seon:PRL2014}.
However, these systems are also restricted to a closed set of target domains and thus cannot provide appropriate responses to \textit{out-of-domain} (OOD) requests.
For example, a dialog system that was designed to cover \texttt{schedule} and \texttt{message} domains could receive OOD requests such as ``\textit{Would you recommend Italian restaurants for me?}'' that is in the \texttt{restaurant} domain or ``\textit{Please record Game of Thrones.}'' that is in the \texttt{TV program} domain.
To maintain user experience, the system must detect OOD requests and provide appropriate back-off responses such as rejection, rather than providing unrelated responses.

The main goal of this paper is to develop an accurate \textit{OOD sentence detection} method.
We define OOD sentence detection as a \textit{binary classification} problem of determining whether the system can respond appropriately to an input sentence, i.e.,
\begin{align}
  f(x) =
    \begin{cases}
    \text{\textit{ID}}, & \text{if $x$ belongs to a domain $d \in D$},\\
    \text{\textit{OOD}}, & \text{otherwise},
  \end{cases}
\end{align}
where $x$ is an input sentence, $D$ is a closed set of target domain-categories such as \texttt{schedule} or \texttt{message}, $ID$ denotes in-domain, and $OOD$ denotes out-of-domain.

Most previous studies~\citep{Nakano:SIGDIAL2011, Tur:Interspeech2014} use both ID sentences and OOD sentences to train OOD sentence detection.
Collecting ID sentences is a necessary step in building many data-driven dialog systems.
However, the task of collecting enough OOD sentences to cover all other domains is laborious and time-consuming.
Therefore, the goal of this paper is to develop an accurate OOD sentence detection method that requires only ID sentences for training.

In this work, we present a novel \textit{neural sentence embedding} method that represents sentences in a low-dimensional continuous vector space that emphasizes aspects that distinguish ID cases from OOD cases.
First, we use large set of unlabeled text to pre-train \textit{word representations} for the initialization of neural sentence embedding.
Second, we use the similarity between OOD sentence detection and \textit{domain-category analysis}~\citep{Komatani:SGIDIAL2009, Li:SLT2014, Nakano:SIGDIAL2011, Tur:ICASSP2012} to train neural sentence embedding with only ID sentences.

\vspace{5pt}
\hspace{0pt}
\hangindent=14.5pt
Domain-category analysis is a task that assigns one of a closed set of target domains to a given sentence; this analysis system can be trained using only ID sentences that are collected for each domain.
We think that the task of OOD sentence detection is more similar to domain-category analysis than to other tasks such sentiment analysis or speech-act analysis, so we expect that the features (i.e., representation) of a sentence extracted by a domain-category analysis system can be used for OOD sentence detection too.

\vspace{5pt}

\noindent Therefore we adopt a feature extractor that is trained for domain-category analysis, and use it as a neural sentence embedding system for OOD sentence detection.
Lastly, the learned representations of ID sentences are used to train an autoencoder that detects whether an input sentence is ID or OOD based on its reconstruction error.
To the best of our knowledge, this is the first study that applies neural sentence embedding to solve the sentence representation problem of OOD sentence detection.

The remainder of this paper is organized as follows:
In Section~\ref{sec:related_work}, we review previous studies.
In Section~\ref{sec:methods}, we describe our method in detail.
In Section~\ref{sec:experiment}, we explain our experimental data, evaluation metrics, and methods to be compared.
In Section~\ref{sec:results}, we show and discuss the experimental results.
In Section~\ref{sec:conclusion}, we conclude this paper.

\section{Related work}
\label{sec:related_work}

Previous studies~\citep{Lane:TASLP2007, Nakano:SIGDIAL2011, Tur:Interspeech2014} on OOD sentence detection use sentence representations based on bag-of-words models, which have limitations in representing rare or unknown words; those words are likely to appear in OOD sentences.
Lane~et~al.~\citep{Lane:TASLP2007} proposed an in-domain verification (IDV) method, which uses only ID sentences to build domain-wise one-vs.-rest classifiers that generate low confidence scores for OOD sentences, and then uses the scores as evidence that a sentence was OOD.
We implemented this method and compared it to our work.
Nakano~et~al.~\citep{Nakano:SIGDIAL2011} proposed an two-stage domain selection framework, which uses both ID sentences and OOD sentences to build multi-domain dialog systems; the main contribution is to use discourse information to prevent erroneous domain switching, but whenever developers expand the domain of a dialog system they must reassess all OOD sentences because some will become ID due to the change of the boundary between ID and OOD.
Tur~et~al.~\citep{Tur:Interspeech2014} used syntactic feature and semantic feature for OOD sentence detection; web search queries are used as OOD sentences to eliminate the need to collect OOD sentences, but such queries are \textit{noisy} because some are actually ID, and they cannot be obtained readily without using a commercial search engine.
Compared to these studies, the main contribution of this paper is a neural sentence embedding method that can understand rare words and unknown words.

Recently, neural sentence embedding methods have been assessed for their ability to solve the sentence representation problem.
Paragraph Vector~\citep{Le:ICML2014} is a well-known method that uses a large set of unlabeled text to learn sentence representations, but the representations are  not optimized for a specific task because they are learned based on \textit{unsupervised} objectives.
In contrast, some researchers have worked on \textit{supervised} sentence embedding particularly for natural language understanding using recurrent neural networks~(RNNs)~\citep{Liu:NIPS2015, Vukotic:Interspeech2015, Yao:Interspeech2013} and long short-term memory (LSTM) networks~\citep{Ravuri:Interspeech2015, Yao:SLT2014, Hakkani-Tur:Interspeech2016}.
However, because we cannot define an objective function based on classification error between ID cases and OOD cases, these methods are not directly applicable to our problem in which only ID sentences are available as a training set.
To solve this problem, we exploit the similarity between OOD sentence detection and domain-category analysis~(Section~\ref{sec:intro}).

Another important part of OOD sentence detection is \mbox{one-class} classification that uses the training data about a target class to distinguish between target items and uninteresting items.
Nearest Neighbor Distances~(NN-d)~\citep{Tax:ICPR2000} classifies an input item as the target class when the local density\footnote{The local density of an item is the distance between the item and its closest item in the training data.} of the item is larger than the local density of its closest item.
A one-class support vector machine~(OSVM)~\citep{Scholkopf:NC2001} learn a decision function about distinguishment.
In this work, we propose to use an autoencoder to detect OOD sentences, and compare the results to those obtained using other methods including NN-d and OSVM.

\section{The proposed OOD-sentence detection method}
\label{sec:methods}

\begin{figure}[t!]
\centering 
\includegraphics[width=0.41\textwidth]{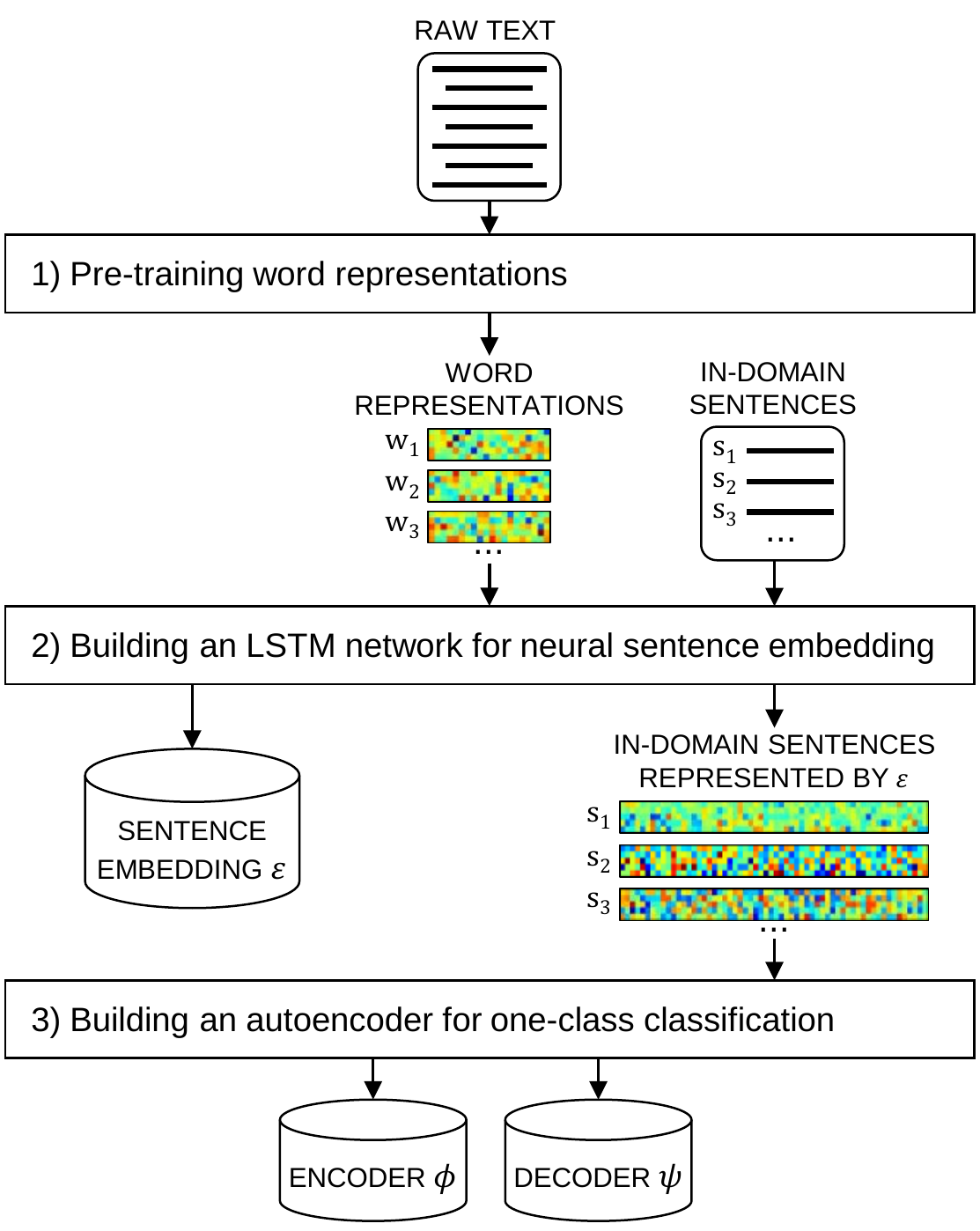}
\caption{Overall training process of our proposing method. Components and processes are described in the text.}
\label{fig:train_flow}
\end{figure}

We defined OOD sentence detection $f(x)$ as a binary classification problem~(Section~\ref{sec:intro}).
However, unlike most other binary classification problems, we assume that only ID sentences are available as training data.
With these ID sentences, domain-category analysis $g(x)=d \in D$ can be built under another assumption that the domain category for each ID sentence is given.

When we represent sentences in $m$-dimensional continuous vector space, we take sentence embedding $\varepsilon(x) \in \mathbb{R}^m$ from an LSTM network trained with $g(x)$ as the supervised objective.
Then, we build an autoencoder that consists of an \textit{encoder} $\phi$ that takes sentences represented by $\varepsilon(x)$ and maps them onto a different space, and a \textit{decoder} $\psi$ that reconstructs their original representations.
Finally, we use the learned autoencoder ($\phi$,~$\psi$) to detect OOD sentences of which reconstruction errors are greater a threshold $\theta$ as:

\begin{align}
  f(x) =
    \begin{cases}
    \text{\textit{ID}}, & \text{if $\lVert\psi(\phi(\varepsilon(x)))-\varepsilon(x)\rVert^2 < \theta$,}\\
    \text{\textit{OOD}}, & \text{otherwise.}
  \end{cases}
\end{align}

The details of the proposed method (Fig.~\ref{fig:train_flow}) are presented in the remainder of this section.

\subsection{Pre-training of word representations}
\label{sec:methods:pre_train}

Before training neural \textit{sentence} embedding, \textit{words} must be represented in a low-dimensional continuous vector space in which semantically-similar words are located near each other.
For example, `\textit{London}' should be closer to `\textit{Paris}' than to `\textit{apple}' in the vector space.
The pre-trained word representations would be \textit{fine-tuned} when sentence representations are learned using ID sentences (Section~\ref{sec:methods:neural_embedding}).
However, the amount of ID sentences is smaller than the amount of unlabeled text such as Wikipedia articles, so pre-training increases both the accuracy and coverage of the word representations.

We utilize the distributional hypothesis of words: the meanings of words can be found by their accompanying words~\citep{Firth:SLA1957}.
We first use a large set of unlabeled texts extracted from Korean Wikipedia articles as the training set, then use it to train a skip\mbox{-}gram neural network~\citep{Mikolov:ICLR2013} that predicts 10 surrounding words by using a $v$-dimensional hidden layer; we set $v$ as 100.
When the training set consists of $k$ unique words in its vocabulary, the result of pre-training is a matrix $\textbf{E} \in \mathbb{R}^{v \times k}$ in which the $i$\textsuperscript{th} column is a $v$-dimensional vector that represents the $i$\textsuperscript{th} word.

\subsection{Neural sentence embedding using an LSTM network}
\label{sec:methods:neural_embedding}

\begin{figure}[t!]
\centering
\includegraphics[width=0.48\textwidth]{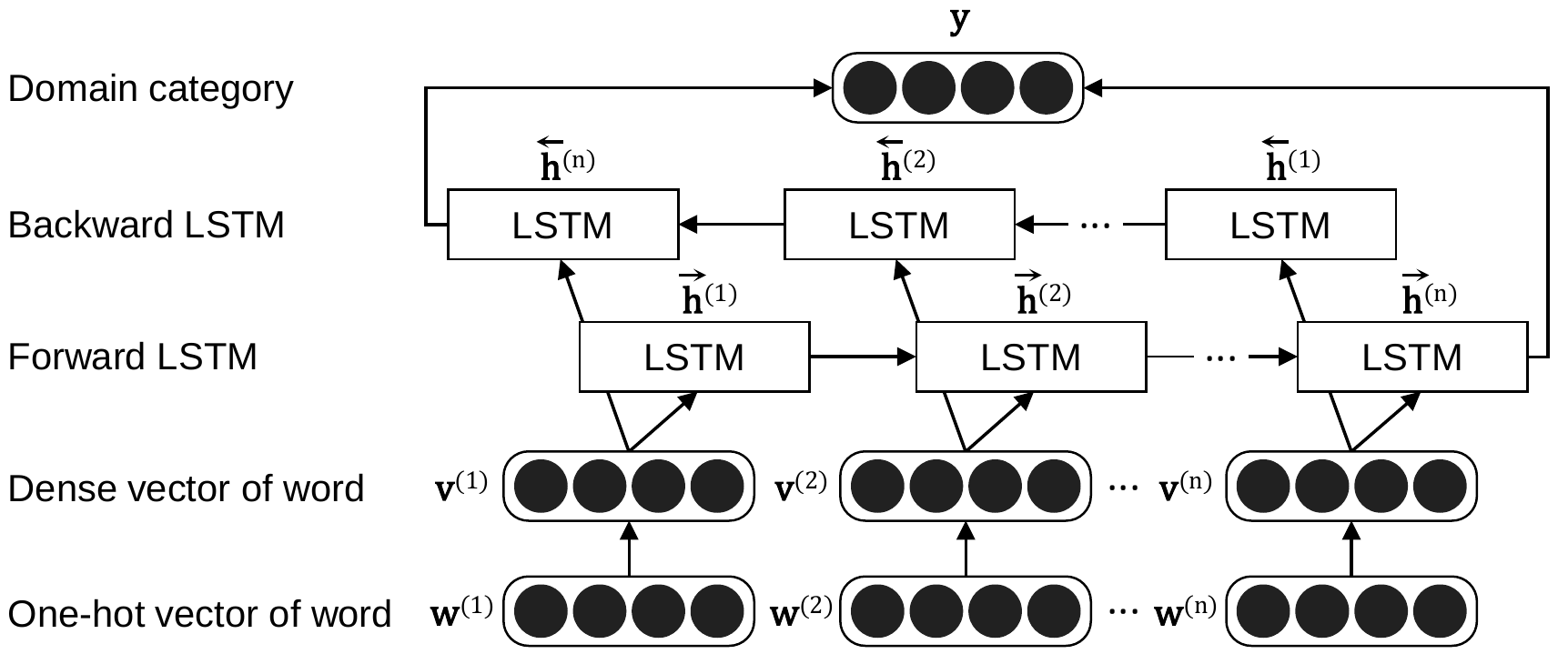}
\caption{LSTM network with two-channel word representations for domain-category analysis. Components and processes are described in the text.}
\label{fig:topic_lstm}
\end{figure}

Sentence embedding~$\varepsilon$ aims to represent given sentences in an $m$-dimensional continuous vector space.
To process variable-length sentences, we use an LSTM network \citep{Hochreiter:NC1997, Gers:NC2000}, which uses a recurrent architecture that learns by repeatedly computing given operations for every word in each sentence.
We suppose that the features of \textit{domain-category analysis} are also important in OOD sentence detection, so we use a set of ID sentences to train a network (Fig.~\ref{fig:topic_lstm}) that classifies a given sentence into a domain category.
The values in the last hidden layer of the trained network represent the given sentence, so $\varepsilon$ can be taken from the trained network.
This is a sort of transfer-learning approach \citep{Pan:KD2010} that learns knowledge from an \textit{auxiliary task} (i.e., domain-category analysis) and applies the knowledge to a \textit{target task} (i.e., OOD sentence detection).

The pre-trained word representations (Section~\ref{sec:methods:pre_train}) are fine-tuned based on back-propagation from the objective function of the LSTM network;
fine-tuning finds \textit{task-specific} word representations, whereas pre-trained word representations describe general word meaning.
However, some words in OOD sentences appear rarely or never in ID sentences; this imbalance hinders the fine-tuning of the word representations, so the word representations cannot be fined-tuned accurately.
To prevent this problem, we use a \textit{two-channel} approach: a \textit{non-static} channel is fine-tuned during training, whereas a \textit{static} channel is fixed.
This multi-channel idea has been used earlier for sentiment analysis \citep{Kim:EMNLP2014} but not for OOD sentence detection.
In addition, we apply \textit{dropout} \citep{Srivastava:JMLR2014} to the non-recurrent layers in our LSTM network.
Dropout is a regularization technique that randomly drops some nodes in artificial neural networks during training.
Especially, dropout prevents our LSTM network from becoming biased toward the non-static channel.

Based on our design, our LSTM network is defined as follows.
Let $\textbf{w}^{(t)} \in \mathbb{N}^k$ be the one-hot vector representation of the $t$\textsuperscript{th} word in a given sentence.
The dense vector representation $\textbf{v}^{(t)} \in \mathbb{R}^{2v}$ of the $t$\textsuperscript{th} word is defined as

\begin{align}
\label{eq:x_t}
\textbf{v}^{(t)}=[{\textbf{E}_{s}} \textbf{w}^{(t)}, {\textbf{E}_{n}} \textbf{w}^{(t)}],
\end{align}
\noindent
where $\textbf{E}_s \in \mathbb{R}^{v \times k}$ is the weight matrix for the static channel and $\textbf{E}_n \in \mathbb{R}^{v \times k}$ is the weight matrix for the non-static channel;
both $\textbf{E}_s$ and $\textbf{E}_n$ are initialized to $\textbf{E}$ that was pre-trained in Section~\ref{sec:methods:pre_train}, but only $\textbf{E}_n$ is fine-tuned during the training; a dropout rate of 50\% is applied to both ${\textbf{E}_{s}} \textbf{w}^{(t)}$ and ${\textbf{E}_{n}} \textbf{w}^{(t)}$.

\begin{figure}[t!]
\centering
\includegraphics[width=0.48\textwidth]{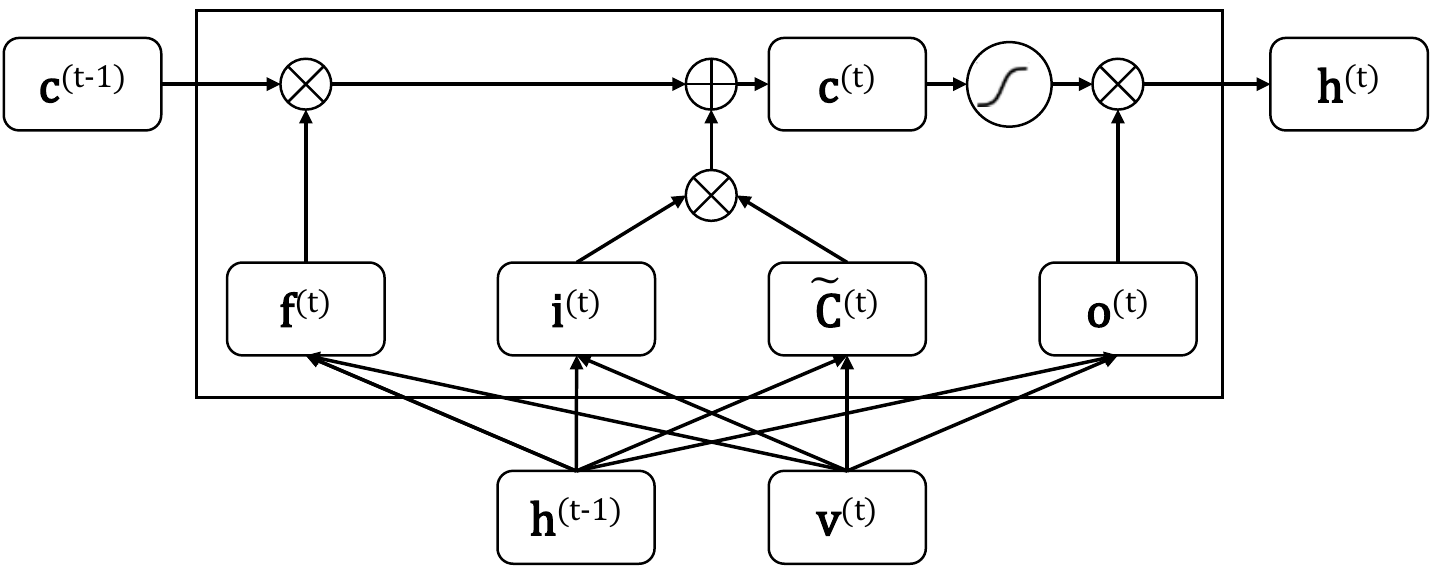}
\caption{LSTM unit. Components and processes are described in the text.}
\label{fig:lstm_unit}
\end{figure}

In $t$\textsuperscript{th} LSTM unit (Fig.~\ref{fig:lstm_unit}), input gate $\textbf{i}^{(t)}$, forget gate $\textbf{f}^{(t)}$, memory cell state $\textbf{c}^{(t)}$, output gate $\textbf{o}^{(t)}$, and hidden state $\textbf{h}^{(t)}$ are defined as
\begin{align}
\textbf{i}^{(t)}&=\sigma(\textbf{W}_i [\textbf{h}^{(t-1)}, \textbf{v}^{(t)}] + \textbf{b}_i), \\
\textbf{f}^{(t)}&=\sigma(\textbf{W}_f [\textbf{h}^{(t-1)}, \textbf{v}^{(t)}] + \textbf{b}_f), \\
\textbf{\~{C}}^{(t)}&=\textrm{tanh}(\textbf{W}_C [\textbf{h}^{(t-1)}, \textbf{v}^{(t)}] + \textbf{b}_C), \\
\textbf{c}^{(t)}&=\textbf{i}^{(t)} \otimes \textbf{\~{C}}^{(t)} + \textbf{f}^{(t)} \otimes \textbf{c}^{(t-1)}, \\
\textbf{o}^{(t)}&=\sigma(\textbf{W}_o [\textbf{h}^{(t-1)}, \textbf{v}^{(t)}] + \textbf{b}_o), \\
\textbf{h}^{(t)}&=\textbf{o}^{(t)} \otimes \textrm{tanh}(\textbf{c}^{(t)}),
\end{align}

\noindent
where $\otimes$ denotes element-wise product, $\sigma$ is the sigmoid activation function, $\textbf{W}_i$, $\textbf{W}_f$, $\textbf{W}_C$, and $\textbf{W}_o$ are weight matrices, and $\textbf{b}_i$, $\textbf{b}_f$, $\textbf{b}_C$, and $\textbf{b}_o$ are bias vectors.
We use bidirectional structure \citep{Graves:NN2005, Schuster:SP1997} of the LSTM network to prevent it from becoming biased toward the last few words, so those weights are defined independently for the forward LSTM and the backward LSTM.

The output domain-category layer $\textbf{y} \in \mathbb{R}^{|D|}$ is computed
based on $[\overrightarrow{\textbf{h}}^{(n)}, \overleftarrow{\textbf{h}}^{(n)}]$, which is the concatenation of the last hidden layers of both the forward LSTM and the backward LSTM, so that
\begin{align}
\textbf{y}=\textrm{softmax}(\textbf{W}_y [\overrightarrow{\textbf{h}}^{(n)}, \overleftarrow{\textbf{h}}^{(n)}] + \textbf{b}_y),
\end{align}
\noindent
where $\textbf{W}_y$ is a weight matrix and $\textbf{b}_y$ is a bias vector; A dropout rate of 50\% is applied to $[\overrightarrow{\textbf{h}}^{(n)}, \overleftarrow{\textbf{h}}^{(n)}]$.

The LSTM network is trained to minimize the categorical cross entropy between the gold standard domain category and the predicted output.
As a result, given a sentence, $[\overrightarrow{\textbf{h}}^{(n)}, \overleftarrow{\textbf{h}}^{(n)}]$ of the trained network represents the sentence in a vector space that emphasizes the distinguishing aspects among domain categories.

\subsection{One-class classification using an autoencoder}
\label{sec:methods:autoencoder}

\begin{figure}[t!]
\centering
\includegraphics[width=0.2\textwidth]{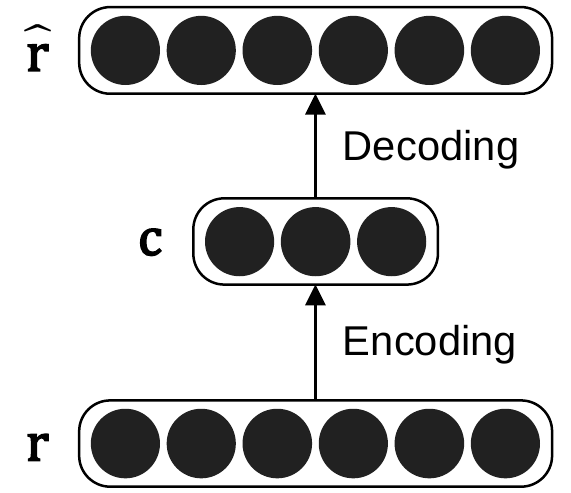}
\caption{Autoencoder.}
\label{fig:autoencoder}
\end{figure}

Autoencoders are feed-forward neural networks that encode and decode given inputs.
When an autoencoder is trained on \textit{interesting} data, the reconstruction error is low for \textit{interesting} input data but high for \textit{uninteresting} input data.
Following the idea of one-class classification \citep{Manevitz:Neurocomputing2007, Xu:COLING2014} based on this characteristic,
we use all ID sentences represented by $\varepsilon$ in Section~\ref{sec:methods:neural_embedding} to train an autoencoder (Fig.~\ref{fig:autoencoder}) that aims to use reconstruction errors to detect OOD sentences.

The autoencoder is the pair of an encoder $\phi$ and a decoder $\psi$.
Let $\textbf{r} \in \mathbb{R}^m$ be a given sentence representation.
Encoding layer $\textbf{c} \in \mathbb{R}^{m/2}$ and decoding layer $\hat{\textbf{r}} \in \mathbb{R}^m$ are defined as
\begin{align}
\textbf{c}&=\textrm{tanh}(\textbf{W}_{\phi} \textbf{r} + \textbf{b}_{\phi}), \\
\hat{\textbf{r}}&=\textrm{tanh}(\textbf{W}_{\psi} \textbf{c} + \textbf{b}_{\psi}),
\end{align}
\noindent
where $\textbf{W}_{\phi}$ and $\textbf{W}_{\psi}$ are weight matrices and $\textbf{b}_{\phi}$ and $\textbf{b}_{\psi}$ are bias vectors.
The autoencoder is trained to minimize the reconstruction error $\lVert\textbf{r} - \hat{\textbf{r}}\rVert^2$.

\section{Experimental setup}
\label{sec:experiment}

\subsection{Implementation details}
\label{sec:experiment:implement}

To implement the pre-training of word representations (Section~\ref{sec:methods:pre_train}), we use Gensim library \citep{Rehurek:LREC2010}; we chose initial learning rate 0.05 and decreased it linearly.
To implement the LSTM network (Section~\ref{sec:methods:neural_embedding}) and the autoencoder (Section~\ref{sec:methods:autoencoder}), we use Keras library \citep{chollet:github2015}; we tried three optimization algorithms (adam \citep{Kingma:ICLR2015}, adadelta \citep{Zeiler:2012}, rmsprop \citep{Tieleman:2012}) for the LSTM network and the autoencoder, and tried two hidden layer sizes (100 and 150) for the LSTM network.
In Section~\ref{sec:results}, we present the result only with the best optimization algorithm and the best hidden layer size, instead of enumerating all results.

\subsection{Data set}
\label{sec:experiment:data_set}

To demonstrate the effectiveness of our proposed method, we experimented on a data set of 7,975 Korean sentences.
The data set was collected manually using a Wizard-of-Oz approach by several groups of people from industries in Korea including LG Electronics and Mediazen.
The data set consists of 5,755 ID sentences for eight domains: \texttt{building guide}, \texttt{car navigation}, \texttt{diet advisor}, \texttt{general}, \texttt{restaurant information}, \texttt{music search}, \texttt{TV program guide}, and \texttt{weather information}; and 2,200 OOD sentences for five domains: \texttt{finance}, \texttt{occupation}, \texttt{small talk}, \texttt{stock}, and \texttt{study}.
Eighty percent of the ID sentences were used to train the models; the remaining ID sentences and all OOD sentences were used for testing.
Although we used Korean sentences to implement our method, it does not rely on language-specific processes except for word tokenization, and can therefore be applied to other languages.

\subsection{Evaluation metrics}
\label{sec:experiment:metrics}

We use equal error rate (EER) to represent the accuracy of OOD sentence detection \citep{Lane:TASLP2007}.
EER is the error rate at which false acceptance rate
\begin{align}
\textnormal{FAR} &= \frac{\textnormal{Number of \textit{accepted} OOD sentences}}{\textnormal{Number of OOD sentences}}
\end{align}

and false rejection rate

\begin{align}
\textnormal{FRR} &= \frac{\textnormal{Number of \textit{rejected} ID sentences}}{\textnormal{Number of ID sentences}}
\end{align}

are equal.

\setlength\tabcolsep{5pt}
\begin{table*}[t!]
\caption{Equal error rates (\%) of OOD sentence detection using combinations of sentence representation methods (row) and classification methods (column). The best result in each sentence representation method (row) is \underline{underlined}; the best result in each classification method (column) is in \textbf{bold}.}
\centering
\begin{tabular}{l l l l l l l}
\toprule
Sentence representation & \multicolumn{6}{l}{Classification} \\
\cmidrule(l){2-7} & NN-d\hspace{42pt} & OSVM\hspace{34pt} & CBC\hspace{42pt} & IDV\hspace{42pt} & Autoencoder & Best \\
\midrule
\multicolumn{7}{l}{\textbf{One-hot encoding:}} \\
BoW							& 26.05 & 29.27 & \underline{11.24} & 13.69 & 21.41 & 11.24 \\
TF-IDF						& 27.62 & 33.78 & \underline{11.00} & 15.83 & 14.00 & 11.00 \\
\midrule
\multicolumn{7}{l}{\textbf{Unsupervised neural sentence embedding:}} \\
Neural BoW					& 27.11 & 28.77 & \underline{20.09} & 26.67 & 34.15 & 20.09 \\
PV-DBOW						& 34.02 & 28.65 & 26.58 & 28.48 & \underline{24.59} & 24.59 \\
PV-DM						& 31.35 & 38.10 & 29.87 & 32.21 & \underline{22.61} & 22.61 \\
\midrule
\multicolumn{7}{l}{\textbf{Supervised neural sentence embedding based on speech-act (SA) analysis:}} \\
SA-RNN w/ random 	& 28.92 & 20.53 & 25.61 & 23.11 & \hphantom{1}\underline{9.18} & \hphantom{1}9.18 \\
SA-RNN w/ static 				& 31.61 & 45.54 & \underline{29.54} & 34.90 & 30.46 & 29.54 \\
SA-RNN w/ non-static			& 27.02 & 26.23 & 29.51 & 26.40 & \underline{18.28} & 18.28 \\
SA-RNN w/ two-channel			& 27.11 & 22.85 & 35.50 & 36.10 & \underline{14.90} & 14.90 \\
SA-LSTM w/ random 	& 27.29 & 22.94 & 38.10 & 16.80 & \hphantom{1}\underline{9.78} & \hphantom{1}9.78 \\
SA-LSTM w/ static				& 27.79 & 35.76 & 25.72 & 35.94 & \underline{12.89} & 12.89 \\
SA-LSTM w/ non-static			& 23.97 & 25.54 & 31.93 & 16.00 & \hphantom{1}\underline{8.50} & \hphantom{1}8.50 \\
SA-LSTM w/ two-channel			& 25.89 & 20.44 & 28.76 & 17.16 & \underline{11.04} & 11.04 \\
\midrule
\multicolumn{7}{l}{\textbf{Supervised neural sentence embedding based on domain-category (DC) analysis:}} \\
DC-RNN w/ random 	& 25.81 & 12.30 & 11.79 & 12.05 & \underline{11.50} & 11.50 \\
DC-RNN w/ static			& 31.68 & 29.69 & 20.27 & 22.25 & \underline{15.52} & 15.52 \\
DC-RNN w/ non-static		& 26.84 & 14.72 & 11.77 & 11.32 & \hphantom{1}\underline{9.16} & \hphantom{1}9.16 \\
DC-RNN w/ two-channel		& 25.63 & 27.44 & 16.36 & 16.38 & \underline{10.75} & 10.75 \\
DC-LSTM w/ random 	& \textbf{19.82} & 15.32 & \textbf{10.73} & \textbf{10.31} & \hphantom{1}\underline{7.44} & \hphantom{1}7.44 \\
DC-LSTM w/ static			& 23.36 & 27.02 & 16.18 & 21.99 & \underline{10.99} & 10.99 \\
DC-LSTM w/ non-static		& 21.21 & 16.27 & 11.77 & 10.57 & \hphantom{1}\underline{7.11} & \hphantom{1}7.11 \\
DC-LSTM w/ two-channel		& 20.27 & \textbf{14.11} & 10.91 & 10.91 & \hphantom{1}\underline{\textbf{7.02}} & \hphantom{1}7.02 \\
\midrule
Best						& 19.82 & 14.11 & 10.73 & 10.31 & \hphantom{1}7.02 & \hphantom{1}7.02 \\
\bottomrule
\end{tabular}
\label{tab:ood_eer}
\end{table*}
\setlength\tabcolsep{6pt}

\subsection{Compared methods}
\label{sec:experiment:baselines}

We evaluated all possible combinations of sentence representation method and classification method.
First, we called our neural sentence embedding method~(Section~\ref{sec:methods:neural_embedding}) \textbf{\mbox{DC-LSTM}} because it uses an LSTM trained for domain category~(DC) analysis.
We compare DC-LSTM to eight sentence-representation methods.
\begin{itemize}[noitemsep, nosep] 

\item \textbf{BoW}. Bag-of-words represents a sentence as a vector in which the $i$\textsuperscript{th} element is the frequency of the $i$\textsuperscript{th} keyword in the sentence. We use $n$-gram by increasing $n$ from 1 to 3 to capture local word order; only the result with the best $n$ is presented in Section~\ref{sec:results}.

\item \textbf{TF-IDF}. A sentence is represented as a vector in which the $i$\textsuperscript{th} element is the product of the term frequency~(TF) and the inverted document frequency~(IDF) of the $i$\textsuperscript{th} keyword in the sentence. We use n-gram as in BoW.

\item \textbf{Neural BoW} \citep{Hermann:ACL2014}. A sentence is represented as the element-wise average of its word representations obtained in Section~\ref{sec:methods:pre_train}.

\item \textbf{PV-DBOW}. This is the distributed BoW version of Paragraph Vector~(Section~\ref{sec:related_work}). We use Doc2Vec implementation in Gensim library \citep{Rehurek:LREC2010}, and set the dimension of sentence representation to 200.

\item \textbf{PV-DM}. This is the distributed memory version of Paragraph Vector~(Section~\ref{sec:related_work}). The rest is the same as PV-DBOW

\item \textbf{SA-RNN}. A sentence is represented by the last hidden layer of the RNN trained for \textit{speech-act} (SA) analysis instead of domain-category analysis.
To do this, we manually annotate speech acts on the same data set; our system includes five speech-act labels: \texttt{question}, \texttt{statement}, \texttt{request}, \texttt{yn-response}, and \texttt{greetings}.
\item \textbf{SA-LSTM}. This is the LSTM network version of SA-RNN.
\item \textbf{DC-RNN}. This is the RNN version of DC-LSTM.
\end{itemize}
In the RNNs and the LSTM networks, we compare four variations of word embedding: \textbf{random}, \textbf{static}, \textbf{non-static}, and \textbf{two-channel}; the first one initializes word embedding randomly, and the others were described in Section \ref{sec:methods:neural_embedding}.

Second, we compare the \textbf{autoencoder} to four classification methods.
\begin{itemize}[noitemsep, nosep]

\item \textbf{NN-d}~(Section~\ref{sec:related_work}). We use the Jaccard distance of two sentences as the distance measure.

\item \textbf{OSVM}~(Section~\ref{sec:related_work}). We use OneClassSVM implementation in scikit-learn library~\citep{Pedregosa:JMLR2011}, and apply three types of kernels to it: linear, polynomial, and radial basis function; only the result with the best kernel is presented in Section~\ref{sec:results}.

\item \textbf{CBC}. The combination of binary classifiers rejects input sentences that are rejected by all the domain-wise one-vs.-rest classifiers.
We use SVC implementation in scikit-learn library \citep{Pedregosa:JMLR2011}, and apply kernels as in OSVM.

\item \textbf{IDV} (Section~\ref{sec:related_work}). In-domain verification (IDV) uses the  individual classifiers as CBC does. However, IDV uses their confidence scores as the feature of classification.

\end{itemize}

\section{Results and discussion}
\label{sec:results}

Our proposed method, autoencoder + DC-LSTM w/ two-channel, was the most accurate~(EER=7.02\%) in the OOD sentence detection experiment (Table~\ref{tab:ood_eer}).
IDV~+~BoW~(EER=13.69\%) is a previous study \citep{Lane:TASLP2007} that used only ID sentences for training.
One-class deep neural network (OCDNN)\footnote{OCDNN uses a \textit{recursive} neural network instead of a \textit{recurrent} neural network.} for opinion-relation detection \citep{Xu:COLING2014} can be applied also to OOD sentence detection, and their method corresponds to autoencoder~+~DC-RNN~w/~non-static (EER=9.16\%).
This result means that our proposed method decreased EER by 23.37\% compared to the state-of-the-art method.
In the remainder of this section, we present five detailed observations from the experiment.

(1) The supervised embedding methods based on \textit{domain-category analysis} were more accurate than the other sentence representation methods such as the supervised embedding methods based on \textit{speech-act analysis}.
This superiority means for neural sentence embedding in OOD sentence detection, domain-category analysis is a more suitable auxiliary task than speech-act analysis.
In contrast, the unsupervised sentence embedding methods cannot optimize the sentence representations for OOD sentence detection, so those methods had higher error rates than even one-hot encoding methods.

\begin{figure}[t!]
\centering 
\includegraphics[width=0.48\textwidth]{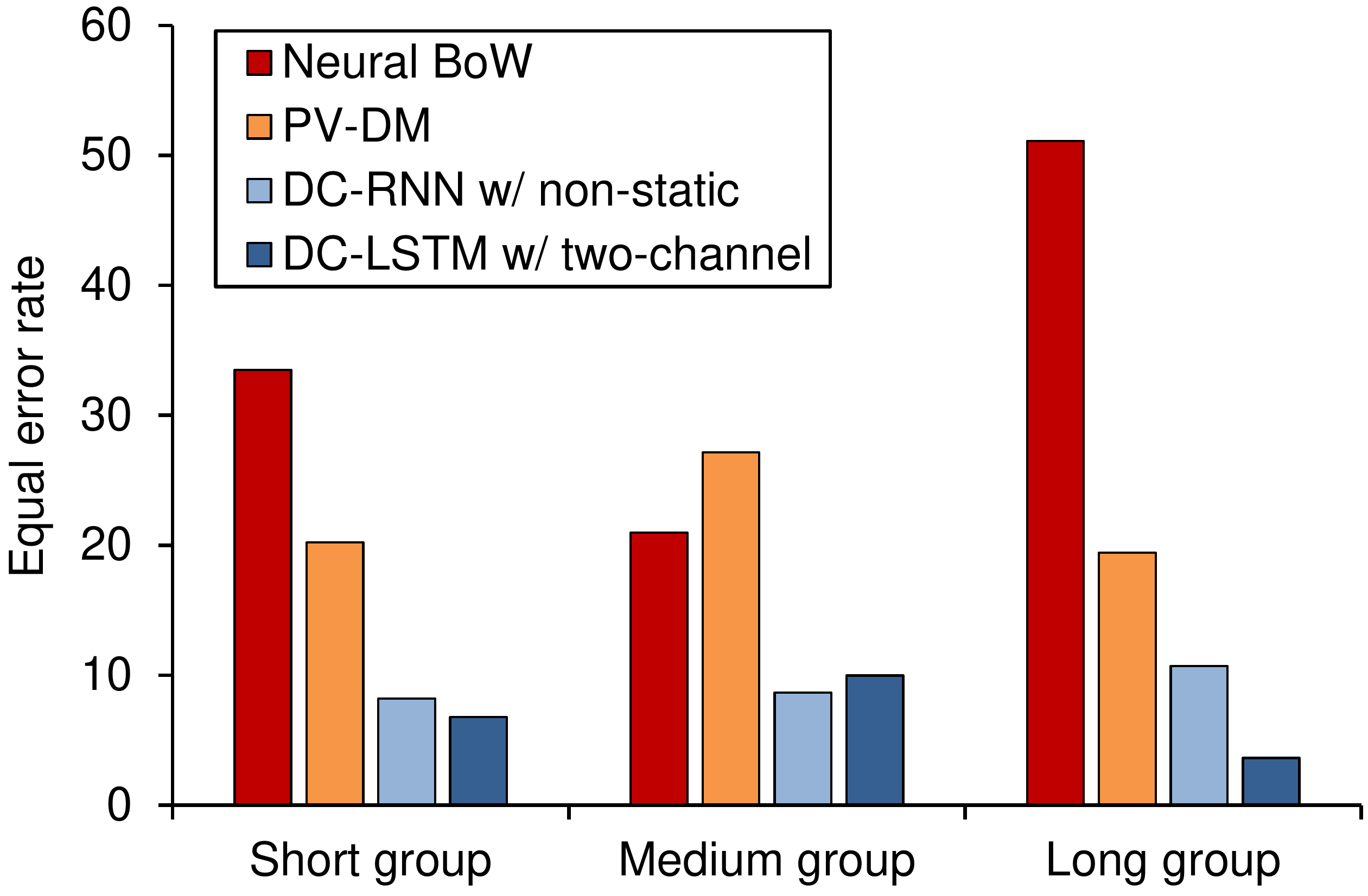}
\caption{Equal error rates (\%) of OOD sentence detection using Neural BoW, PV-DM, DC-RNN with non-static word embedding, and DC-LSTM with two-channel word representations. The autoencoder is used for classification. Sentences in \texttt{short} (31.05\% of sentences), \texttt{medium} (38.00\%), and \texttt{long} group (30.96\%) had 1-8, 9-11, and 12-22 words respectively.}
\label{fig:ood_eer_by_len}
\end{figure}

(2) DC-LSTMs were more accurate than DC-RNNs.
To compare them in detail, we divided the test set into \texttt{short}, \texttt{medium}, and \texttt{long} groups based on the length of each sentence (Fig.~\ref{fig:ood_eer_by_len});
DC-LSTM w/ two-channel greatly reduced the error rate of DC-RNN w/ non-static in the \texttt{long} group (-7.08\% points), although the difference was small in the \texttt{short} group (-1.43\% points) and the \texttt{medium} group (+1.33\% points).
This result means that, in OOD sentence detection, LSTM networks reduce the vanishing-gradient problem of standard RNNs.

\begin{table}[t!]
\caption{Accuracies (\%) of domain-category analysis.}
\centering
\begin{tabular}{l l l}
\toprule
Method						& Accuracy \\
\midrule
SVM + BoW			& 95.50 \\
SVM + TF-IDF		& 95.93 \\
RNN w/ random				& 96.19 \\
RNN w/ static				& 93.16 \\
RNN w/ non-static			& 96.28 \\
RNN w/ two-channel			& 96.28 \\
LSTM w/ random				& 96.02 \\
LSTM w/ static				& 96.45 \\
LSTM w/ non-static			& \textbf{96.80} \\
LSTM w/ two-channel			& 96.37 \\
\bottomrule
\end{tabular}
\label{tab:dca_acc}
\end{table}

(3) DC-LSTM w/ two-channel was more accurate than DC-LSTM w/ random, w/ static, or w/ non-static.
Nevertheless, the best accuracy of \textit{domain-category analysis} itself was achieved by LSTM w/ non-static rather than by LSTM~w/~two-channel~(Table~\ref{tab:dca_acc}); we think the reason is that the accuracy of domain-category analysis can be increased by fine-tuning the representations of \textit{only} known words.
To summarize, the two-channel approach is effective in OOD sentence detection, but we cannot say that it is effective in general.

\begin{figure}[t!]
\centering
\includegraphics[width=0.37\textwidth]{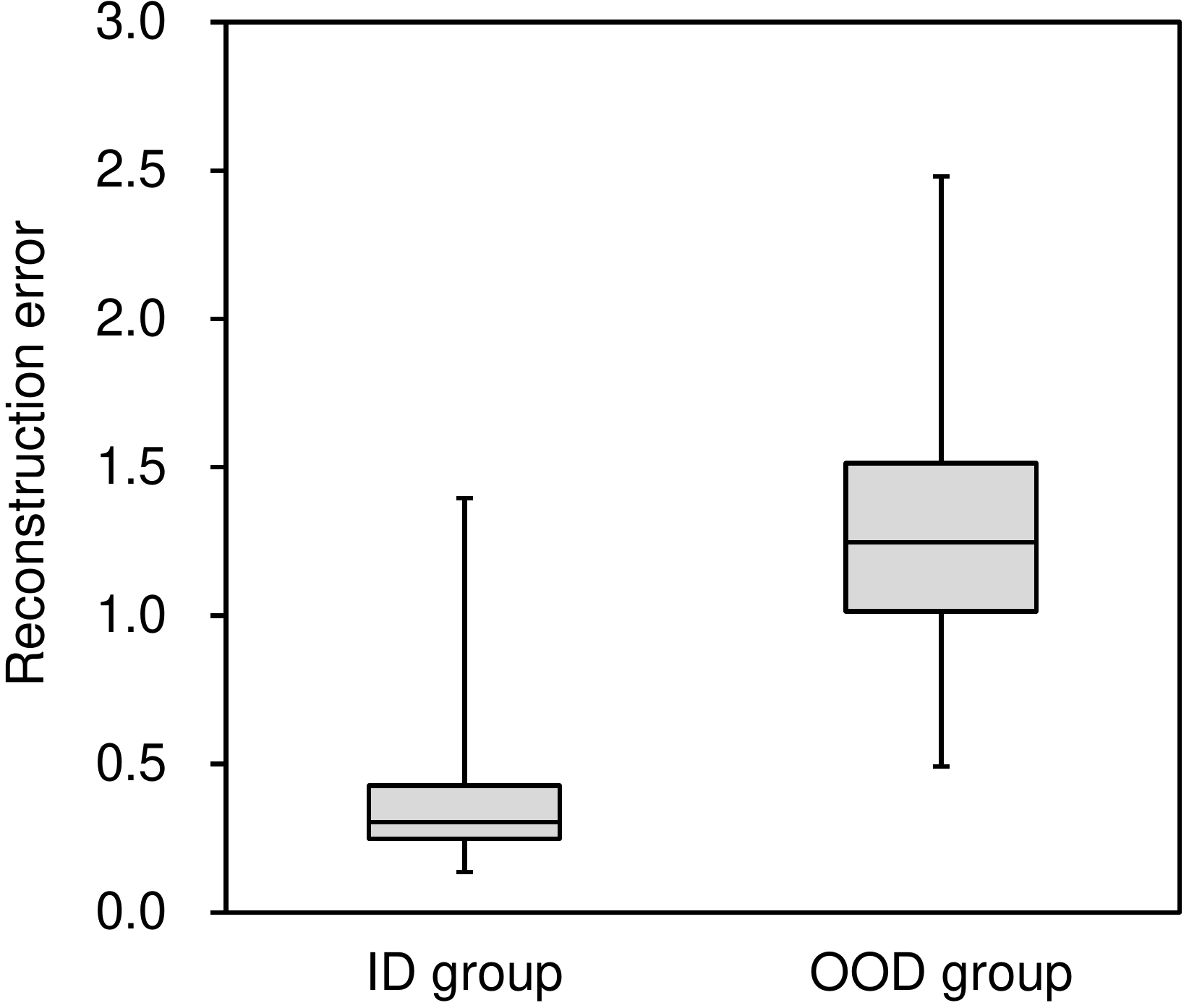}
\caption{Reconstruction errors by autoencoder + DC-LSTM w/ two-channel.}
\label{fig:ood_score}
\end{figure}

\begin{figure}[t!]
\centering
	\begin{subfigure}[t]{0.235\textwidth}
		\includegraphics[width=\textwidth]{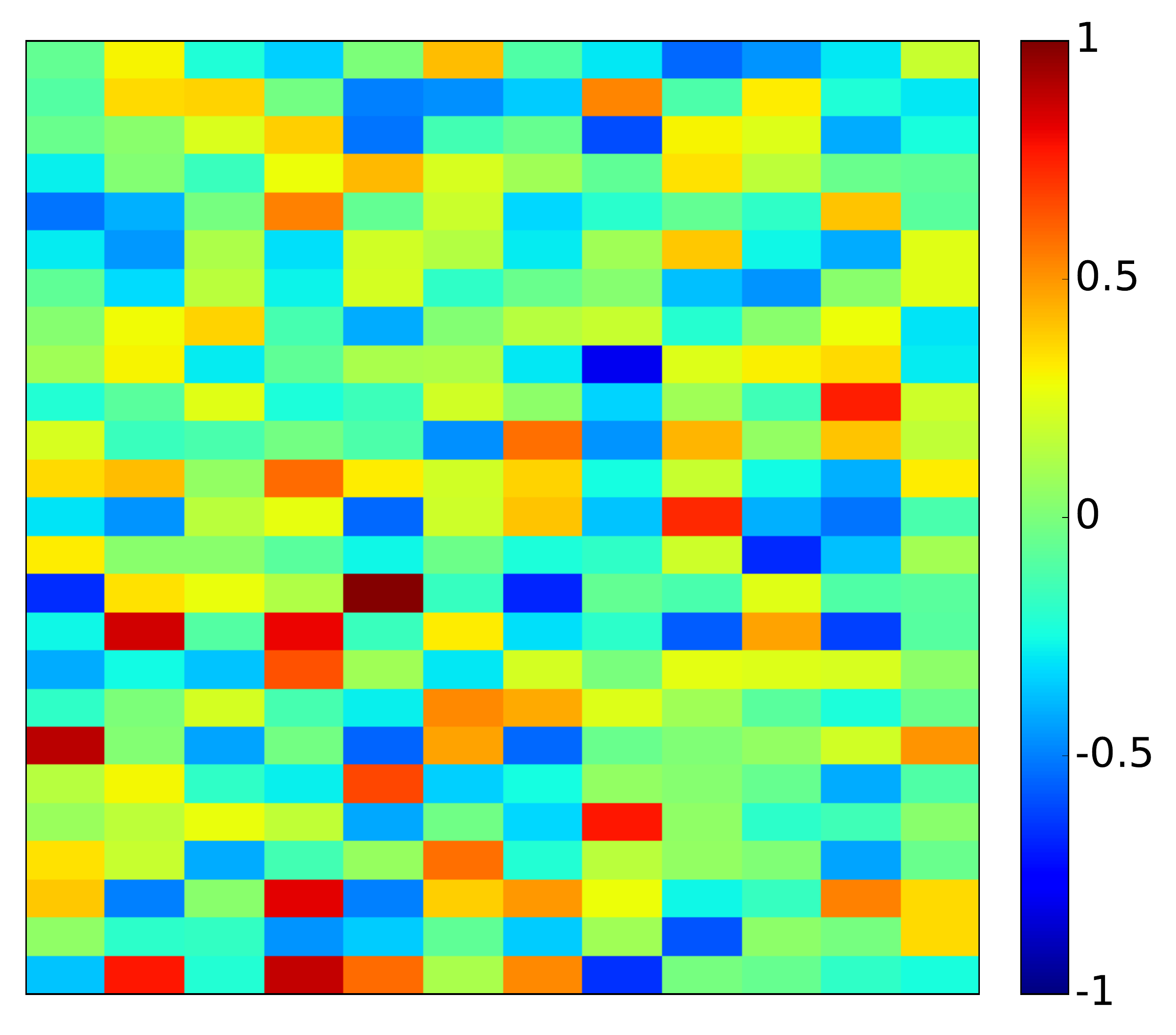}
		\caption{Autoencoder input}
		\label{fig:autoencoder_example_ood_input}
	\end{subfigure}
	\begin{subfigure}[t]{0.235\textwidth}
		\includegraphics[width=\textwidth]{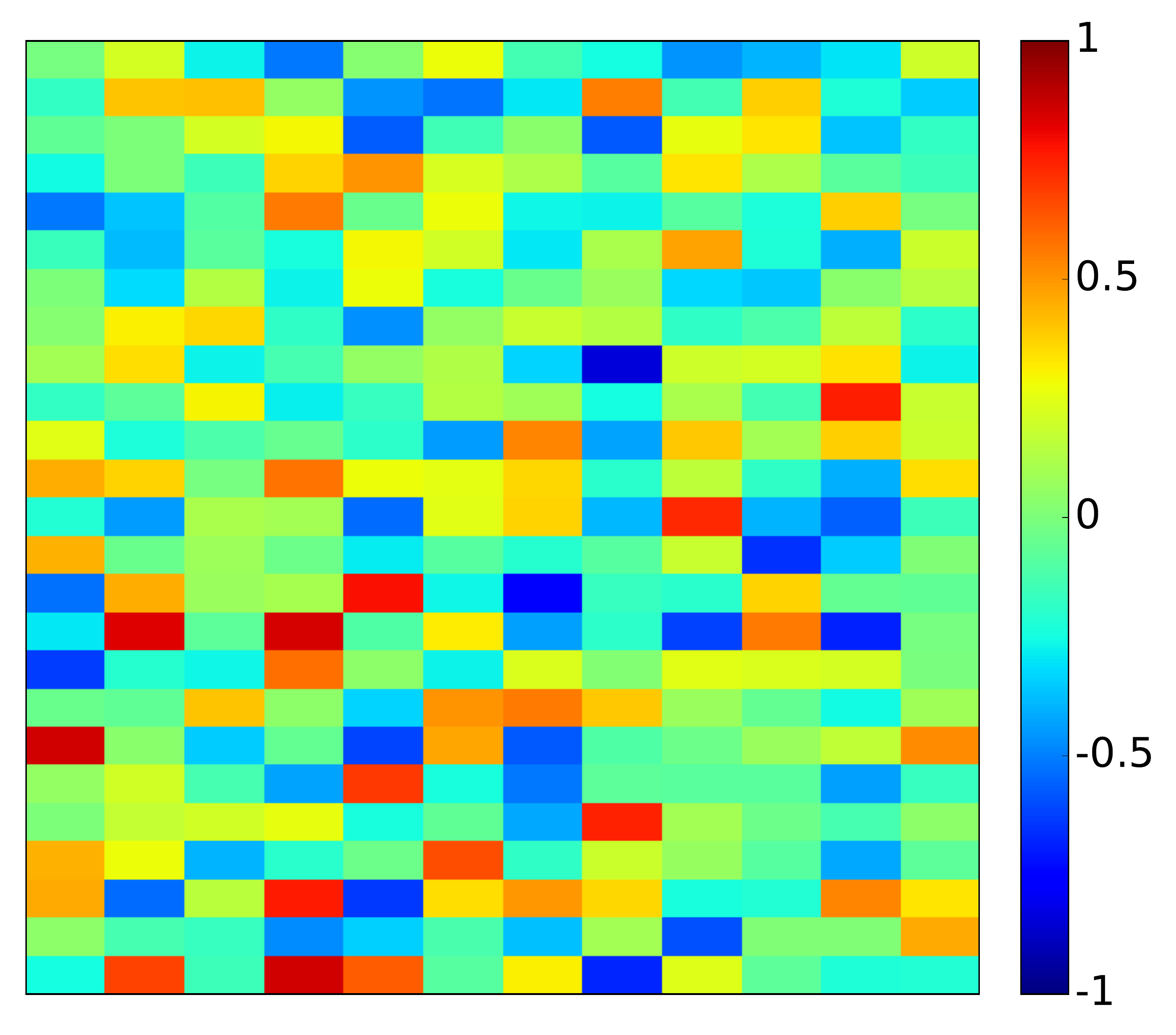}
		\caption{Autoencoder output}
		\label{fig:autoencoder_example_ood_output}
	\end{subfigure}
\caption{(a) the representation of an OOD sentence "\textit{I think that investment cannot be learned in a day}" by DC-LSTM w/ two-channel and (b) its reconstruction by the autoencoder. We plot 300-dimensional representation vectors of which real-valued cells ranges from -1.0 to 1.0 as 25-by-12 matrices.
The difference (i.e., reconstruction error) between (a) and (b) was 1.57, although it is subtle to the naked eye.}
\label{fig:autoencoder_example_ood}
\end{figure}

(4) The autoencoder was the best classification method for DC-LSTM w/ two-channel.
As expected, the reconstruction errors in the autoencoder were low for ID sentences but high for OOD sentences on average (Fig.~\ref{fig:ood_score} and Fig.~\ref{fig:autoencoder_example_ood}).
This result means that the reconstruction error by the autoencoder is reliable evidence that a sentence is OOD.

\begin{figure}[t!]
\centering
\includegraphics[width=0.48\textwidth]{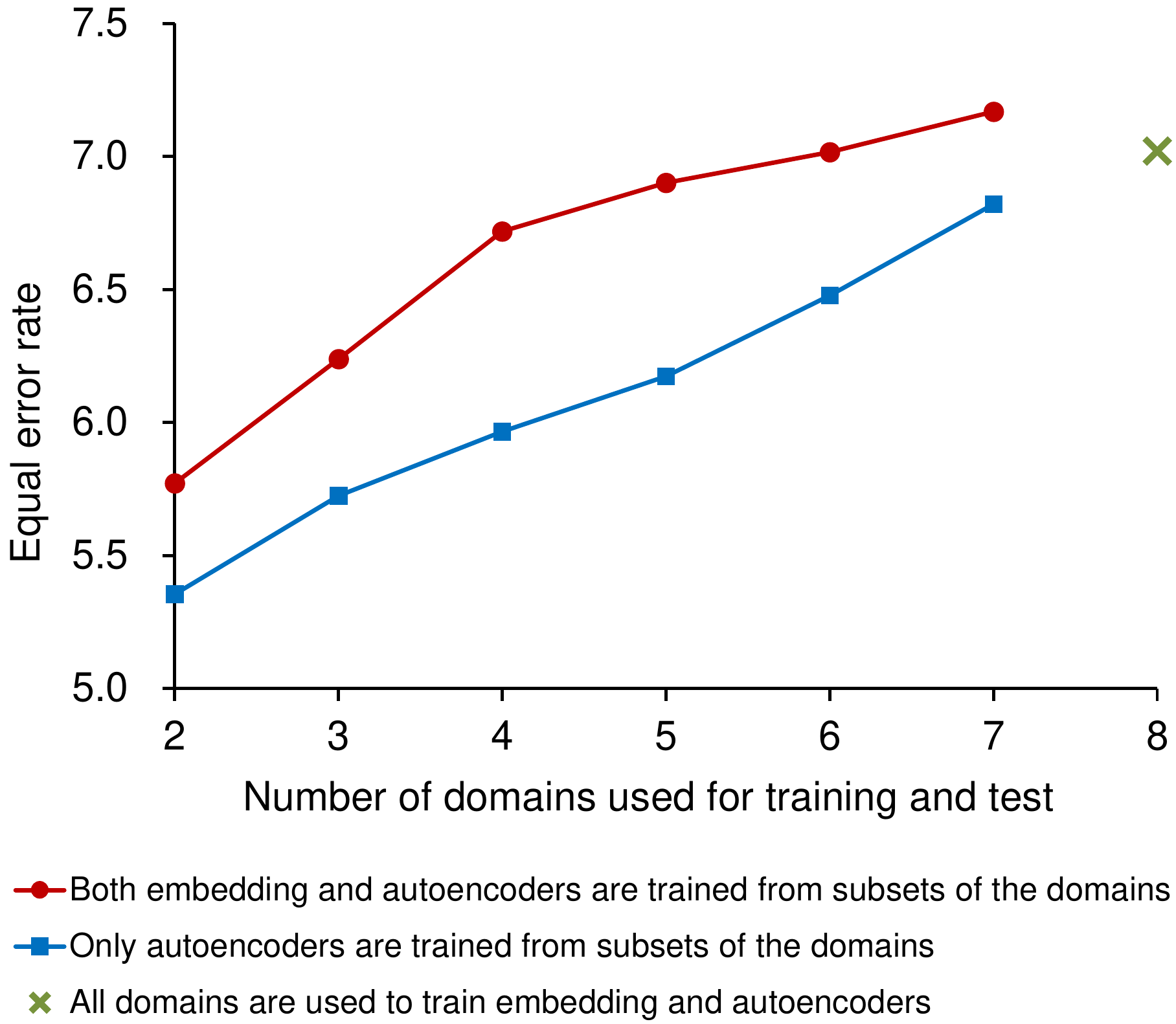}
\caption{Equal error rates (\%) of OOD sentence detection in which the number of domains ranges from 2 to 8.}
\label{fig:domain_number}
\end{figure}

(5) The number of domains affected the result.
When we used \textit{subsets} of the domains by varying the number of target domains from 2 to 7 to train and test OOD sentence detection, the average EER was proportional to the number of domains (red circles in Fig~\ref{fig:domain_number}) although the EER decreased when the number of domains was increased from 7 to 8 (green cross in Fig~\ref{fig:domain_number}).
However, the results were improved when all domains were used to train sentence embedding (blue squares in Fig~\ref{fig:domain_number}).
Therefore, we can conclude that increasing the number of domains can decrease the accuracy of autoencoders but can increase the accuracy of sentence embedding.

\section{Conclusion}
\label{sec:conclusion}

To detect OOD sentences, we developed a method that is trained using only ID sentences.
We used an LSTM network trained for domain-category analysis as a neural sentence embedding system for OOD sentence detection because the features for domain-category analysis are also effective for OOD sentence detection; the word representations were pre-trained using a large set of unlabeled text before domain-category analysis was trained.
We used the learned sentence representations to train an autoencoder that detects OOD sentences based on their reconstruction errors.
In an experiment on a data set of an eight-domain dialog system, the proposed method achieved higher accuracy than the state-of-the-art methods.
This method will help to improve user experience in dialog systems by enabling them to detect OOD sentences accurately.

\section*{Acknowledgments}

This research was supported by the MSIP, Korea, under the G-ITRC support program (IITP-2016-R6812-16-0001) supervised by the IITP.
We thank the editor and two anonymous reviewers for their insightful comments.
We also thank Dr.~Derek~Jon~Lactin for his careful proofreading.

\bibliographystyle{model2-names}
\bibliography{\jobname}

\end{document}